\documentclass[10pt,twocolumn,letterpaper]{article} 

\usepackage{avss}
\usepackage{times}
\usepackage{epsfig}
\usepackage{graphicx}
\usepackage{amsmath}
\usepackage{amssymb}

\avssfinalcopy % *** Uncomment this line for the final submission
\def\avssPaperID{148} % *** Enter the AVSS Paper ID here

\ifavssfinal\fi

\begin{document}

%%%%%%%%% TITLE
\title{Online Tracking Parameter Adaptation based on Evaluation}

\author{Duc Phu Chau \and Julien Badie \and Fran\c cois Br\'emond \and Monique Thonnat \\
STARS team, INRIA, France \\
2004 route des Lucioles, 06560 Valbonne, France\\
{\tt\small \{Duc-Phu.Chau, Julien.Badie, Francois.Bremond, Monique.Thonnat\}  @inria.fr}
% \and
% {\small\url{http://www.author.org/~second}}
}

\maketitle
% \thispagestyle{empty}

%%%%%%%%% ABSTRACT
\begin{abstract}
  Parameter tuning is a common issue for many tracking algorithms. In order to solve this problem, this paper proposes an online parameter tuning to adapt a tracking algorithm to various scene contexts. In an offline training phase, this approach learns how to tune the tracker parameters to cope with different contexts. In the online control phase, once the tracking quality is evaluated as not good enough, the proposed approach computes the current context and tunes the tracking parameters using the learned values. The experimental results show that the proposed approach improves the performance of the tracking algorithm and outperforms recent state of the art trackers. This paper brings two contributions: (1) an online tracking evaluation, and (2) a method to adapt online tracking parameters to scene contexts.
    
%   Object tracking quality usually depends on video scene contexts (\eg scene illumination, object occlusion level). In order to overcome this limitation, we propose in this paper an automatic control approach which is able 
  
\end{abstract}

%%%%%%%%% BODY TEXT
\section{Introduction}

Many studies have been proposed to track the movements of objects in a scene \cite{yilmaz, dpchauVisapp11, benfold11}. However the selection of a tracking algorithm for an unknown scene becomes a hard task. Even when the tracker has already been determined, it is difficult to tune its parameters to get the best performance due to the variations of scene context  (\eg scene illumination, object occlusion level, 2D object sizes).

Some approaches have been proposed to address these issues. The authors in \cite{kuo10} propose an online learning scheme based on Adaboost to compute a discriminative appearance model for each mobile object. However the online Adaboost process is time consuming. The author in \cite{hall} proposes two strategies to regulate the parameters for improving the tracking quality. In the first strategy, the parameter values are determined using an enumerative search. In the second strategy, a genetic algorithm is used to search for the best parameter values. This approach does not require human supervision and parameter knowledge for controlling its tracker. However, it is computationally expensive because of the parameter optimization stage performed in the online phase.

% For example, the authors in \cite{kuo10} propose an algorithm for learning a discriminative appearance model for different mobile objects. First, the object tracklets are determined based on three object descriptors: position, size and color histogram in consecutive frames. Second, training samples are collected online from object tracklets within a time sliding window based on some spatial-temporal constraints. An Adaboost-based learning process is performed to find the most discriminative object local descriptors. This approach has two disadvantages: (1) the object tracklet determination is not reliable because of using simple object descriptors, and (2) the online Adaboost process is time consuming.

In the other hand, some approaches integrate different trackers and then select the convenient tracker depending on video content. For example, the authors in \cite{prost, yoon12} present tracking frameworks which are able to control a set of different trackers to get the best performance. The system runs the tracking algorithms in parallel. At each frame, the best tracker is selected to compute the object trajectories. These two approaches require the execution of different trackers in parallel which is expensive in terms of processing time. In \cite{dpchauIcdp11}, the authors propose a tracking algorithm whose parameters can be learned offline for each tracking context. However the authors suppose that the context within a video sequence is fixed over time. Moreover, the tracking context is selected manually.

These studies have obtained relevant results but show strong limitations on the online processing time and the self-adaptation capacity to the scene variations. In order to solve these problems, we propose in this paper a new method to adapt the tracking algorithms to the scene variations. The principle of the proposed approach is the automatic parameter tuning of tracking algorithms over time during the online process. This parameter tuning relies on an offline learning process and an online tracking evaluation method. The proposed tracking evaluation is responsible for detecting the tracking errors and activating the parameter tuning process if necessary. The parameter tuning relies entirely on the offline learned database, this helps to avoid slowing down the processing time of the tracking task. The variation of scene over time during the online phase is also addressed in the proposed approach.

This paper is organized as follows. Sections 2 and 3 present in detail the proposed approach. Section 4 shows the results of the experimentation and validation. A conclusion as well as future work are presented in section 5.

% -----------------------------------------
\section{Offline Learning}

\begin{figure*}[]
\center
\includegraphics[width=0.96\linewidth] {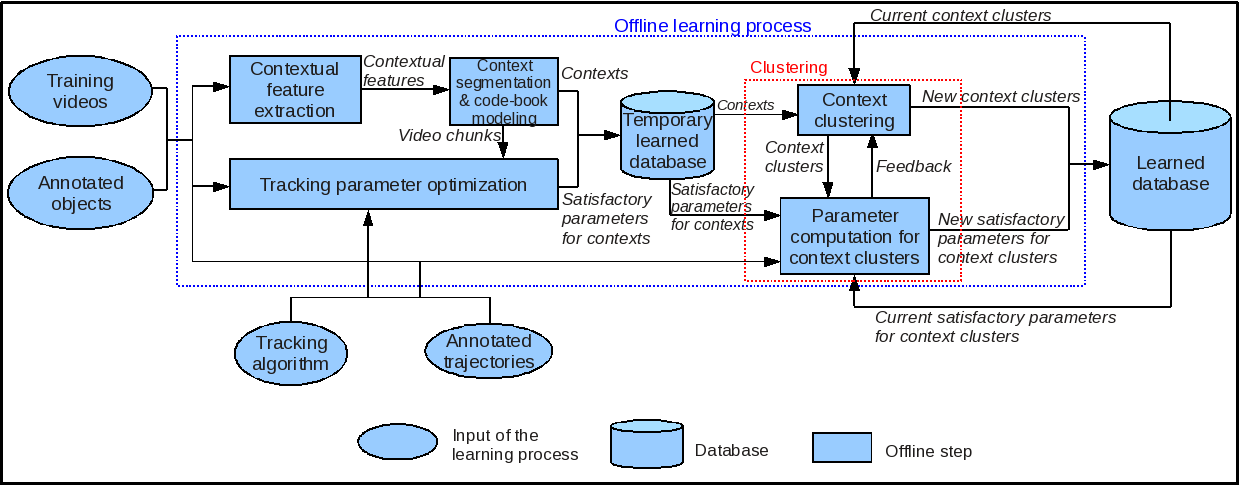} 
\caption{The offline learning scheme}
\label{fig_learning_scheme}
\end{figure*}

The objective of the learning phase is to create a database which supports the control process of a tracking algorithm. This database contains satisfactory parameter values of the tracking algorithm for various contexts. This phase takes as input training video sequences, annotated objects, annotated trajectories, a tracking algorithm including its control parameters. The term ``control parameters'' refers to parameters which are considered in the control process (i.e. to look for satisfactory values in the learning phase and to be tuned in the online phase). At the end of the learning phase, a learned database is created. A learning session can process many video sequences. Figure \ref{fig_learning_scheme} presents the proposed scheme for building the learned database.

The notion of ``context'' (or ``tracking context'') in this work represents elements in the videos which influence the tracking quality. More precisely, a context of a video sequence is defined as a set of six features: density of mobile objects, their occlusion level, their contrast with regard to the surrounding background, their contrast variance, their 2D area and their 2D area variance. The offline learning is performed as follows.

First, for each training video, the \textbf{``contextual feature extraction''} step computes the contextual feature values from annotated objects for all video frames. 

Second, in the \textbf{``context segmentation and code-book modeling''} step, these context feature values are used to segment the training video in a set of consecutive chunks. Each video chunk has a stable context. The context of a video chunk is represented by a set of six code-books (corresponding to six context features). 

Third, the \textbf{``tracking parameter optimization''} is performed to determine satisfactory tracking parameter values for the video chunks using annotated trajectories. These parameter values and the set of code-books are then inserted into a temporary learned database.

%  TODO co the giam 2 cai nay thanh 1 nhugn phai sua lai hinh
After processing all training videos as three above steps, a \textbf{''clustering''} step, which is composed of two sub-steps \textbf{``context clustering''} and \textbf{``parameter computation for context clusters''}, is performed. In the first sub-step, the contexts are clustered using a QT clustering. In the second one, for each context cluster, its satisfactory tracking parameter values are defined in function of the tracking parameters learned for each element context. The context clusters and their satisfactory tracking parameters are then inserted into the learned database.

% -----------------------------------------
\section{Online Control}

\begin{figure}[]
\center
\includegraphics[width=8.65cm] {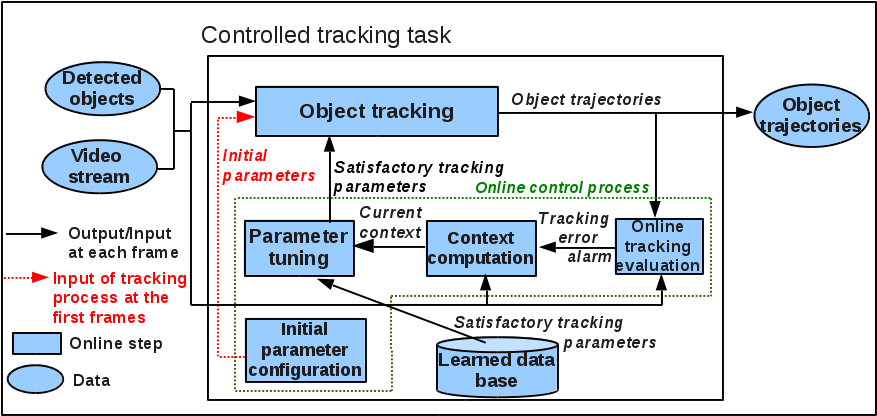} 
\caption{The online control}
\label{fig_online_control}
\end{figure}

In this section, we present in detail how the tracking algorithm is controlled to adapt itself to the contextual variations. This controller takes as input the video stream, the list of detected objects at every frame, the offline learned database, the object trajectories and gives as output the satisfactory tracking parameter values to parameterize the tracker if necessary (see figure \ref{fig_online_control}). 

At each frame, the tracking quality is estimated online. When a tracking error is detected, the proposed controller computes the context of the $n$ latest frames. This context is then used for finding the best matching context cluster in the offline learned database. If such a context cluster is found, the tracking parameters associated with this context cluster are used. In the following sections, we describe the three steps of this phase: online tracking evaluation, context computation and parameter tuning.

% TODO  co the dua flow chart neu co khong gian

\subsection{Online Tracking Evaluation}

%  tranh dung tu framework, module, confidence hay risk, viec chia module la ko tot, va that ra chi la 1

In this paper, we propose a method to estimate online the tracking quality. The main advantage of this approach is that it can be used for evaluating any tracking algorithm. This method takes as input the current object trajectories, the processing video stream, and gives as output at each frame an alarm of tracking quality if necessary.

The principle of this evaluation method relies on the following hypothesis: a tracked object is supposed to have a coherence (\ie low variation) on some appropriate descriptors. The selection of appropriate object descriptors is crucial. These descriptors have to satisfy three criteria: they have to be representative of the tracked object, discriminative enough for distinguishing with the other objects, and can take into account the popular tracking errors (\eg ID switch, ID lost). Regarding these criteria, we use the following five descriptors to evaluate the tracking quality of a mobile object: 2D bounding box, speed, direction, color histogram and color covariance \cite{dpchauIcdp11}. 

Using these descriptors, we define two values representing the tracking risks for each object at each frame. The first one is the object interaction score which takes into account the occlusion and density between the consider object and its spatial neighbors. The second value is called ``object tracking error score'' that evaluates the variations of the last four above object descriptors over time. A high tracking error score (near to 1) alerts a tracking problem such as ID switch or ID lost. In the following sections, we present in detail how to compute these two scores.

% of objects  is an overlap score representing the possibility of interaction between objects (\eg spatial overlap, cross each other) in the scene. The higher the overlap score, the higher the risk of detection and tracking is. This score is computed using the spatial overlap features.

% The second value is the detection risk score. This score is based on a weighted sum of the dimension variance and the status of the object. A high confidence score (near to 1) can be used to detect a merge or a slit of two objects or a clutter.

\subsubsection{Object Interaction Score}

The object interaction score is computed at every frame and for each object. It represents the interaction possibility between mobile objects (\eg spatial overlap, cross each other). This score takes into account the density of mobile objects at the current instant and the object occlusion levels in the last two frames. 

Given an object at instant $t$, denoted $o_t^i$, we can find its neighbors, denoted $\mathfrak{N} (o_t^i)$, which are the spatially close objects. The density score for the object $o_t^i$ is defined as follows:
\begin{equation}
 d (o_t^i) = \frac{union (o_t^i, \mathfrak{N} (o_t^i)) } { cover (o_t^i, \mathfrak{N} (o_t^i)) }
\end{equation}

\noindent where $union(o_t^i, \mathfrak{N} (o_t^i))$ is the union of 2D areas occupied by object $o_t^i$ and its neighbors $\mathfrak{N} (o_t^i)$; $cover (o_t^i, \mathfrak{N} (o_t^i))$ is the area of the smallest rectangular which covers $o_t^i$ and $\mathfrak{N} (o_t^i)$.

%  TODO ve hinh minh hoa neu co space

In order to compute the occlusion level of an object, we define first the occlusion level between two objects $o_t^i$ and $o_t^j$ as follows:

% Given two objects $i$, $j$ at instant $t$ of respectively 2D areas $a_t^i$ and $a_t^j$, we compute their occlusion level based on their area overlap as follows:
\begin{equation}
\mathcal{O} (o_t^i, o_t^j )\ =\ \frac {a_t^{ij}} {min (a_t^i, a_t^j)}
\end{equation}

\noindent where $a_t^i$ is 2D area of object $i$ at time $t$, $a_t^{ij}$ is the overlap area of objects $i$ and $j$ at $t$. 

Second, the occlusion level between object $o_t^i$ and its neighbors, denoted $\mathcal{O}_t (o_t^i)$, is defined as the $max \{  \mathcal{O} (o_t^i, o_t^j ) \ | \ j \in \mathfrak{N} (o_t^i) \}  $. In the same way, we compute the occlusion level between object $o_t^i$  and its neighbors detected at $t-1$, denoted $\mathcal{O}_{t-1} (o_t^i)$.

The interaction score of the object $o_t^i$, denoted $I(o_t^i)$, is defined as the mean value of its density score and the two occlusion level scores $\mathcal{O}_{t-1} (o_t^i)$, $\mathcal{O}_{t} (o_t^i)$:
\begin{equation}
 I(o_t^i) = \frac{d(o_t^i) + \mathcal{O}_{t-1} (o_t^i) + \mathcal{O}_t (o_t^i) }{3}
\end{equation}

\subsubsection{Object Tracking Error Score}

The object tracking error score is computed at every frame and for each object. It represents the potential error on the tracking quality of the considered tracked object. This scores takes into account the variations of the four object descriptors: object speed, direction, histogram color and color covariance over time. The 2D bounding box descriptor is not used because it is very dependent on the detection quality. For each object descriptor at instant $t$, we compute the mean and standard deviation values, denoted $\mu_t^k$ and $\delta_t^k$, where $k$ representing the considered descriptor ($k = 1..4$). The tracking error score of an object at $t$ is defined as follows:
\begin{equation}
E_t = \frac{ \sum_{\alpha=1}^{4} \frac {\delta_t^\alpha} {\mu_t^\alpha}  } { 4 } 
 \end{equation}

 \subsubsection{Object Tracking Error Alarm}
 \label{sec_error_alarm}
 
At instant $t$, a tracked object is considered as ``erroneous'' if its interaction score and tracking error score are greater than a same threshold $Th_1$; and its tracking error score increases by a predefined threshold $Th_2$ compared to its tracking error score computed at $t-1$. If there exists such a tracked object, the tracking evaluation task sends a tracking error alarm to the context computation task to improve the tracking performance.

\subsection{Context Computation}
\label{sec_context_computation}

The context computation task is only activated when the tracker fails. The objective of this step is to find the context cluster stored in the offline learned database to which the context of the current processing video belongs. This step takes as input for every frame, the list of the current detected objects and the processing video stream. First, we compute the six context feature values (density, occlusion level, contrast, contrast variance, 2D area and 2D area variance of mobile objects) of the video chunk corresponding to the last $n$ frames ($n$ is a predefined parameter). The set of these feature values is denoted $\mathfrak{C}$. Second, let $\mathfrak{D}$ represent the offline learned database, a context feature set $\mathfrak{C}$ belongs to a cluster $C_i$ if both conditions are satisfied:

\begin{equation}
\label{eq_cond_belonging_cluster_thrld}
contextDistance(\mathfrak{C},\ C_i)\ <\ Th_3 \\
\end{equation}
\begin{equation}
\label{eq_cond_belonging_cluster_min}
\begin{array}{llr}
\forall C_j\ \in \ \mathfrak{D}, j \neq i: \\
\ contextDistance(\mathfrak{C}, C_i) \leq  contextDistance(\mathfrak{C}, C_j) 
\end{array}
\end{equation}
	
\noindent where $Th_3$ is a predefined threshold; $contextDistance(\mathfrak{C},\ C_i)$ represents the distance between a context feature set $\mathfrak{C}$ and a context cluster $C_i$. This distance relies on the number of times where the context feature values belonging to $\mathfrak{C}$ matches to code-words in $C_i$. 

\subsection{Parameter Tuning}

If such a context cluster $C_i$ is found, the satisfactory tracking parameters associated with $C_i$ are used for parameterizing the tracking of the current video chunk. Otherwise, the tracking algorithm parameters do not change, the current video chunk is marked to be learned offline later.

%------------------------------------------------------------------------ 
\section{Experimental Results}

%  caretaker, pets ( fan tich de cai thien kq) minh hoa 1 so hinh anh ket qua, tud,   caviar (lay kq tu icvs13)  
%  xem 1 so paper moi de so sanh kq (avss 12)
%  co thoi gian thi test them 1 so kq moi tu cac bai cua cua ecc12 dc copy trong thu muc tud
% kiem tra lai gt cua tud-sad chi co 6 nguoi, xem bao cua aandriey cvpr12 de them 1 so papers de so sanh

\subsection{Parameter Setting and Object Detection Algorithm}

The proposed control method has four predefined parameters. The first two parameters are thresholds $Th_1$ and $Th_2$, presented at section \ref{sec_error_alarm}, are respectively set to 0.2 and 0.15. The third parameter is the distance threshold $Th_3$ (section \ref{sec_context_computation}) is set to $0.5$. The last parameter is the number of frames $n$ to compute the context, presented at section \ref{sec_context_computation}, is set to $50$. These parameter values are unchanged for all the experiments presented in this paper. A HOG-based algorithm \cite{corvee10} is used for detecting people in videos.

\subsection{Tracking Evaluation Metrics}

In this experimentation, we use the following tracking evaluation metrics. Let $GT$ be the number of trajectories in the ground-truth of the test video. The first metric \textbf{$MT$} computes the number of trajectories successfully tracked for more than 80\% divided by GT. The second metric \textbf{$PT$} computes the number of trajectories that are tracked between 20\% and 80\% divided by GT. The last metric \textbf{$ML$} is the percentage of the left trajectories.

\subsection{Controlled Tracker}
\label{sec_controlled_tracker}
In this paper, we select an object appearance-based tracker \cite{dpchauIcdp11} to test the proposed approach. This tracker takes as input a video stream and a list of objects detected in a predefined temporal window. The object trajectory computation is based on a weighted combination of five object descriptor similarities: 2D shape ratio, 2D area, RGB color histogram, color covariance and dominant color. For this tracker, the five object descriptor weights $w_k$ ($k$ = 1..5, corresponding to the five above descriptors) are selected for testing the proposed control method. These parameters depend on the tracking context and have a significant effect on the tracking quality.

\subsection{Training Phase}
\label{sec_tracker_1_training_phase}

In the training phase, we use 15 video sequences belonging to different context types 
(i.e. different levels of density and occlusion of mobile objects as well as of their contrast with regard to the surrounding background, their contrast variance, their 2D area and their 2D area variance). These videos belong to four public datasets (ETISEO, Caviar, Gerhome  and PETS), to two European projects (Caretaker and Vanaheim). They are recorded in various places: shopping center, buildings, home, subway stations and outdoor.

Each training video is segmented automatically in a set of context segments. Each object descriptor similarity can be considered as a weak classifier for linking two objects detected within a temporal window. Therefore in the tracking parameter optimization process, we use an Adaboost algorithm to learn the object descriptor weights for each context segment. The Adaboost algorithm has a lower complexity than the other heuristic optimization algorithms (e.g. genetic algorithm, particle swam optimization). Also, this algorithm avoids converging to the local optimal solutions. After segmenting the 15 training videos, we obtain 72 contexts. By applying the clustering process, 29 context clusters are created.

\subsection{Testing Phase}

All the following test videos do not belong to the set of the 15 training videos.

\subsubsection{Subway video}

\begin{figure*}[]
\begin{center}$
\begin{array}{ccc}
\includegraphics[width=5cm, height = 3cm]{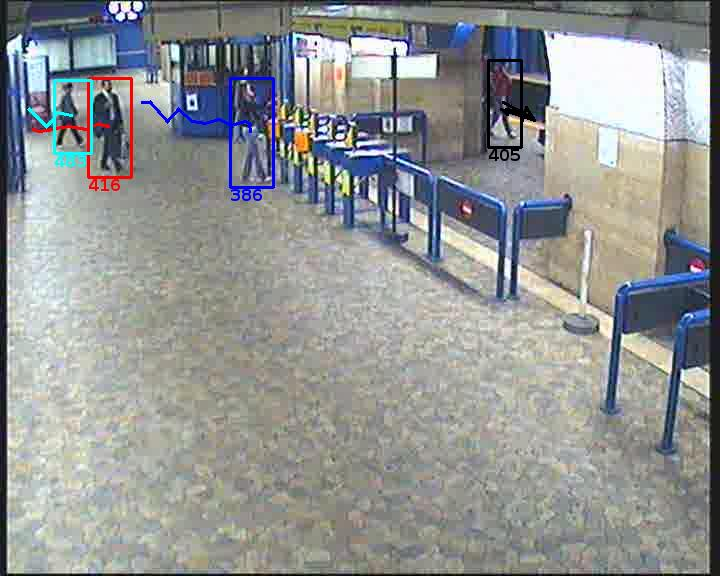} &
\includegraphics[width=5cm, height = 3cm]{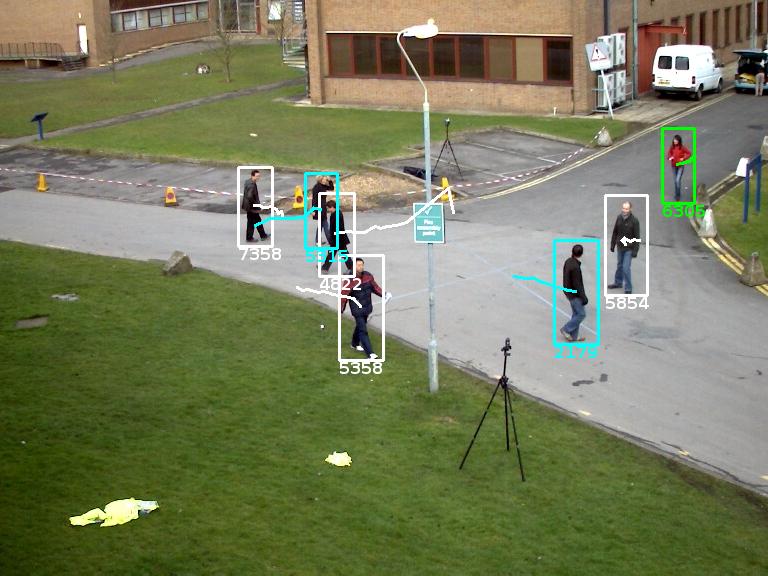} &
\includegraphics[width=5cm, height = 3cm]{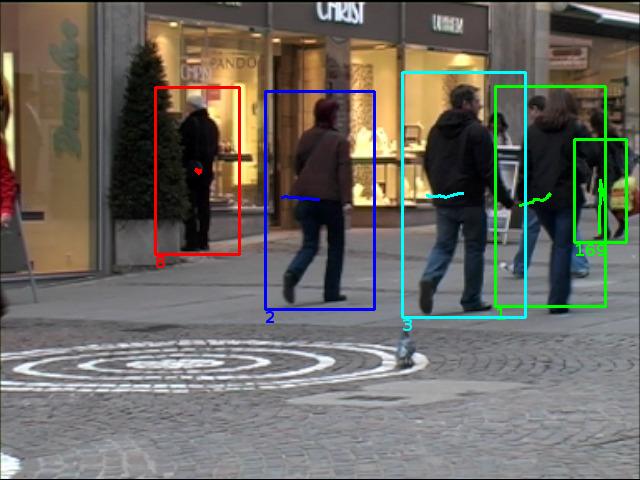} \\
\end{array}$
\end{center}
\caption{\label{fig_dataset} Illustration of the output of the controlled tracking process for three videos: Left image: Subway video; Middle image: PETS 2009 S2L1, time 12:34; Right image: TUD-Stadtmitte. Different IDs represent different tracked objects. The object trajectories are only displayed for the last 10 frames.}
\end{figure*}

The first tested video sequence belongs to the Caretaker European project whose video camera is installed in a subway station (see the left image of the figure \ref{fig_dataset}). The length of this sequence is 5 minutes. It contains 38 mobile objects.

% \begin{figure}[b]
% \centering
% \includegraphics[width=0.6\linewidth]{img_caretaker/caretaker_frame_468.png}
% \caption{Illustration of the tracking result for the subway video. Different IDs represent different tracked objects. The object trajectories are only displayed for the last 10 frames}
% \label{fig_illustration_caretaker}
% \end{figure}

\begin{figure*}[]
\begin{center}$
\begin{array}{ccc}
\includegraphics[width=5cm, height=2.5cm]{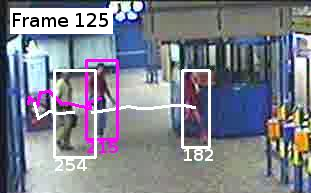} &
\includegraphics[width=5cm, height=2.5cm]{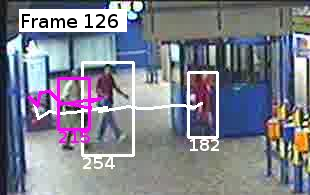} &
\includegraphics[width=5cm, height=2.5cm]{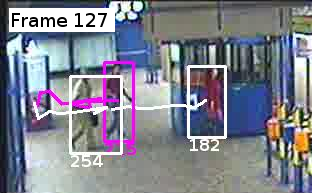} \\
\end{array}$
\end{center}
\caption{\label{fig_caretaker}Illustration of the output of the controlled tracking process. Different IDs represent different tracked objects.}
\end{figure*}

Figure \ref{fig_caretaker} illustrates the output of the controlled tracking process. We consider the tracking result of the two persons on the left images. At the frame 125, these two persons with respectively ID 254 (the left person) and ID 215 (the right person) are correctly tracked. Person 254 has a larger bounding box than person 215. At the frame 126, due to an incorrect detection, the left person has a quite small bounding box. By consequence, the IDs of these two persons are switched because the tracking algorithm currently uses object 2D area as an important descriptor. Now the online tracking evaluation sends an alarm on tracking error to the context computation task. The context cluster associated to the following parameters are selected for tuning the tracking parameters: $w_1 = 0$, $w_2 = 0$, $w_3 = 0.72$, $w_4 = 0$ and $w_5 = 0.28$ (see section \ref{sec_controlled_tracker} for the meaning of these parameters). The color histogram which is selected now as the most important descriptor ($w_3 = 0.72$). The 2D area descriptor is not used ($w_2 = 0$). At the frame 127, after the tracking parameter tuning, the two considered objects take the correct IDs as in frame 125.

% The 2D areas of these objects are very small. The color histogram descriptor which takes into account all pixels belonging to the objects, can better discriminate mobile objects than the dominant color descriptor. The color covariance descriptor is neither not reliable for objects of low resolution.

Table \ref{tab_caretaker_result} presents the tracking results of the tracker \cite{dpchauIcdp11} in two cases: without and with the proposed controller. We find that the proposed controller helps to improve significantly the tracking performance. The value of $MT$ increases 52.7\% to 84.2\% and the value of $ML$ decreases 18.4\% to 10.5\%.

\begin{table}[b]
   \begin{center}
	\begin{tabular}{|p{3.6 cm}|p{1 cm}|p{1 cm}|p{1 cm}|}
% 	\begin{tabular}{|l|c|c|c|c|}
		\hline
			  Methods 		& MT(\%) & PT(\%) & ML(\%)   \\
		\hline
Chau et al. \cite{dpchauIcdp11} without the proposed controller & 52.7 	& 28.9 & 18.4	\\
		\hline
\textbf{Chau et al. \cite{dpchauIcdp11} with the proposed controller} &  \textcolor{red} {\textbf{84.2}} & 5.3 & \textcolor{red} {\textbf{10.5}}\\
% \textbf{Chau et al. \cite{dpchauIcdp11} with the proposed controller} &  \textcolor{red} {\textbf{84.21}} & 5.26 & 10.53\\
		\hline
	\end{tabular}
\end{center}
\caption{\label{tab_caretaker_result}Tracking results of the subway video. The proposed controller improves significantly the tracking performance. The best values are printed in \textcolor{red} {\textbf{red}}.}
\end{table}

\subsubsection{PETS 2009 Dataset}

In this test, we select the sequence S2\_L1, camera view 1, time 12.34 belonging to the PETS 2009 dataset for testing because this sequence is experimented in several state of the art trackers (see the middle image of the figure \ref{fig_dataset}). This sequence has 794 frames, contains 21 mobile objects and several occlusion cases. In this test, we use the CLEAR MOT metrics presented in \cite{petsmetric} to compare with other tracking algorithms. The first metric is MOTA which computes Multiple Object Tracking Accuracy. The second metric is MOTP computing Multiple Object Tracking Precision. We also define a third metric $\overline{M}$ representing the average value of MOTA and MOTP. All these metrics are normalized in the interval $[0,\ 1]$. The higher these metrics, the better the tracking quality is.

For this sequence, the tracking error alarms are sent six times to the context computation task. For all these six times, the context cluster associated to the following tracking parameters is selected for tracking objects: $w_1 = 0$, $w_2 = 0.14$, $w_3 = 0.12$, $w_4 = 0.13$ and $w_5 = 0.61$ (see section \ref{sec_controlled_tracker} for the meaning of these parameters). The dominant color descriptor ($w_5$) is selected as the most important descriptor for tracking objects. This selection is reasonable. This descriptor can well handle the object occlusion cases (see \cite{dpchauIcdp11} for more details) which happen frequently in this video. Table \ref{tab_pets_result} presents the metric results of the proposed approach and four recent trackers from the state of the art. With the proposed controller, the tracking result increases significantly. We also obtain the best values in all the three metrics.
\begin{table}[]
   \begin{center}
	\begin{tabular}{|p{3.6 cm}|p{1 cm}|p{1 cm}|p{1 cm}|}
% 	\begin{tabular}{|l|c|c|c|}
		\hline
			  Methods 	& MOTA & MOTP & $\overline{M}$   \\
 		\hline
	
	 	Berclaz et al. \cite{berclaz11}  	& 0.80	& 0.58 	& 0.69	\\
		\hline
		 Shitrit et al. \cite{shitrit11}  	& 0.81	& 0.58 	& 0.70	\\
		\hline
% 		 \cite{..}  	& 0.81	& 0.74 	& 0.78	\\
% 		\hline
		 Henriques et al. \cite{henriques11}  	& \textbf{\textcolor{red}{0.85}}	& 0.69 	& 0.77	\\
		\hline
	      Chau et al. \cite{dpchauIcdp11} without the proposed controller & 0.62 	&0.63  &0.63	\\
		\hline
\textbf{Chau et al. \cite{dpchauIcdp11} with the proposed controller}	& \textbf{\textcolor{red}{0.85}} & \textbf{\textcolor{red}{0.71}} & \textbf{\textcolor{red}{0.78}}  \\
		\hline
	\end{tabular}
\end{center}
\caption{\label{tab_pets_result}Tracking results on the PETS sequence S2.L1, camera view 1, time 12.34. The best values are printed in \textcolor{red} {\textbf{red}}.}
\end{table}

%  TODO them 1 so hinh minh hoa ket qua tracker
\begin{table}[b]
   \begin{center}
% 	\begin{tabular}{|p{10 cm}|p{1.4 cm}|p{1.4 cm}|p{1.4 cm}|}
	 \begin{tabular}{|p{3.6 cm}|p{1 cm}|p{1 cm}|p{1 cm}|}
% 	\begin{tabular}{|p{3 cm}|c|c|c|c|}
		\hline
			  Methods 		& MT(\%) & PT(\%) & ML(\%)   \\
		\hline
			  Kuo et al.	 \cite{kuo11}		& 60 	& 30.0 		& \textbf{\textcolor{red}{10.0}}  	\\
	      \hline
Andriyenko et al. \cite{andriyenko11}		& 60.0 	& 30.0 		& \textbf{\textcolor{red}{10.0}}  	\\
		\hline
Chau et al. \cite{dpchauIcdp11} without the proposed controller	& 50.0 & 30.0 & 20.0 \\
		\hline
		\textbf{Chau et al. \cite{dpchauIcdp11} with the proposed controller}		& \textbf{\textcolor{red}{70.0}} & \textbf{10.0} & \textbf{20.0} \\
		\hline
	\end{tabular}
\end{center}
\caption{\label{tab_tud_result}Tracking results for the TUD-Stadtmitte sequence. The best values are printed in \textcolor{red} {\textbf{red}}.}
\end{table}

\subsubsection{TUD Dataset}

For the TUD dataset, we select the TUD-Stadtmitte sequence. This video contains only 179 frames and 10 objects but is very challenging due to heavy and frequent object occlusions (see the right image of the figure \ref{fig_dataset}). Table \ref{tab_tud_result} presents the tracking results of the proposed approach and three recent trackers from the state of the art. We obtain the best $MT$ value compared to these two trackers.
% \begin{figure}[]
% \begin{center}
%    \includegraphics[width=0.6\linewidth]{img_tud/tud_frame_26.jpg}
% \end{center}
%    \caption{Illustration of object tracking for the TUD-Stadtmitte video. Different IDs represent different tracked objects. The object trajectories are only displayed for the last 10 frames.}
% \label{fig_tud}
% \end{figure}

\subsection{Computational Cost}

All experiments presented in this paper have been performed in a machine of Intel(R) CPU @ 2.60GHz and 8GB RAM. The average processing time of the tracking process for all test videos is 13 fps while using the proposed controller, and is 15 fps without the controller. We find that the controller increases only slightly the computational cost.

% In all the testing video sequences, the online processing time increases only slightly (less than 10\%) when the proposed controller is used.

%------------------------------------------------------------------------ 
\section{Conclusion and Future Work}

In this paper, we have presented a new control approach to adapt the tracking performance to various tracking context. While using the proposed online tracking evaluation, tracking errors are detected quickly. The parameter tuning is then activated to improve immediately the tracking quality. The experiments show a significant improvement of the tracking performance when the proposed controller is used. Although we only address the parameter tuning problem, the proposed approach can also be applied to select online trackers to adapt better the context variations. In future work, the tracking parameters will be learned by an unsupervised method to remove completely the human knowledge from training phase. 

% to adapt the tracking performance to the ``unknown'' tracking context of the test video.

\section*{Acknowledgments}
\noindent This work is supported by The European Vanaheim, Panorama and Support projects.

{\small
\bibliographystyle{ieee}
\bibliography{avss_refs}
}

\end{document}